\documentclass[12pt]{article}
\usepackage{openwork}

\usepackage{url}
\usepackage{amsmath}

\usepackage{varwidth}

\makeatletter
\renewcommand{\fnum@table}{\small\textbf{\tablename~\thetable}}
\renewcommand{\fnum@figure}{\small\textbf{\figurename~\thefigure}}
\newcommand{\fixedfirstpagefootnote}[1]{%
  \begingroup
  \renewcommand\thefootnote{}%
  \renewcommand\@makefnmark{}%
  \long\def\@makefntext##1{\noindent ##1}%
  \footnotetext{#1}%
  \addtocounter{footnote}{0}%
  \endgroup
}
\makeatother

\title{Exploring Geographic Relative Space in Large Language Models through Activation Patching}
\author[*]{Stef De Sabbata}
\author[*] {Rahul Baiju}
\author[**]{Stefano Mizzaro}
\author[**] {Kevin Roitero}
\affil[*]{School of Geography, Geology and the Environment, University of Leicester, UK}
\affil[**]{Department of Mathematics, Computer Science and Physics, University of Udine, Italy}
\date{}

\begin{document}

\maketitle
\fixedfirstpagefootnote{Published in “Proceedings of the 1st International Conference on Geospatial Artificial Intelligence (GeoAI 2026) – Oral Presentation Papers”, edited by Haosheng Huang and Nico Van de Weghe, GeoAI 2026, 3-6 June 2026, Ghent, Belgium. \newline\newline
This contribution underwent single-blind peer review based on the extended abstract.
}

\begin{abstract}
The increased use of Large Language Models (LLMs) in geography raises substantial questions about the safety of integrating these tools across a wide range of processes and analyses, given our very limited understanding of their inner workings. In this extended abstract, we examine how LLMs process relative geographic space using activation patching, an emerging tool for mechanistic interpretability. 
\end{abstract}

\section{1. Introduction}

Mechanistic interpretability \parencite[see e.g.,][]{lindsey2025biology} aims to understand the inner workings of Large Language Models (LLMs). As the use of LLMs expands within geography \parencite[see e.g.,][]{li2025giscienceeraartificialintelligence}, it becomes necessary to develop specialised studies that can provide grounds for the safe deployment of LLMs within the discipline, and provide us with a better understanding of how LLMs handle geographic concepts and knowledge. \textcite{desabbata2025geomechinterp} have demonstrated how LLMs process geographic information spatially. However, approaches such as probing and sparse autoencoders can only provide preliminary evidence \parencite{smith2025negative}. Activation patching \parencite{NEURIPS2020_92650b2e} has recently gained prominence as a method that can demonstrate causal relationships between activations and identify reasoning patterns.

This extended abstract applies activation patching to investigate how LLMs process relative space, a fundamental concept in quantitative geography \parencite{o2024computing}.

\section{2. Activation patching}

Activation patching \parencite{NEURIPS2020_92650b2e} was introduced as an approach designed to identify the locations of specific concepts or knowledge within the parameters of an LLM. 
The central principle of activation patching is to intentionally disrupt information processing at a specific point within the model. If the disruption at a specific point does not affect the model's output for a given prompt, that point likely contributes minimally (if at all) to processing that prompt. However, if the disruption alters the output, the disruption point is relevant to the query and the associated concepts or knowledge.

Current approaches typically use a procedure with two prompts and three forward passes \parencite{zhang2024bestpracticesactivationpatching}. Using the terminology commonly used in activation patching, first, a \textit{clean} prompt is processed by the model, capturing all \textit{clean} activations and the output probabilities for the most probable next token. Second, a \textit{corrupted} prompt is processed, and its \textit{corrupted} output probabilities are recorded. Third, the \textit{corrupted} prompt is processed again, but activations from the \textit{clean} prompt are injected at a specific computation point, thus \textit{patching} the computation by replacing the ones generated by the \textit{corrupted} prompt. The resulting \textit{patched} outputs are compared to the \textit{corrupted} outputs to assess whether the patching point is related to the concepts or knowledge present in the prompt.

\section{3. Methodology}

To investigate how LLMs process relative space, this study explores how Gemma 2 2B\footnote{https://huggingface.co/google/gemma-2-2b} processes the concept of \textit{near} \parencite{Derungs01102016}. To that end, we use a set of \textit{clean} prompts in the format shown below, where \texttt{<placename>} is one of 249 placenames referring to UK populated places with over fifty thousand people according to the GeoNames free gazetteer dataset\footnote{https://download.geonames.org/export/dump/}. Each \textit{clean} prompt is then used to \textit{patch} a set of distance-focused \textit{corrupted} prompts in the format shown below where \texttt{<placename>} matches the corresponding \textit{clean} prompt and \texttt{<distance>} is one of the following quantitative expressions of distance: ``\texttt{five miles}'', from ``\texttt{ten miles}'' to ``\texttt{a hundred miles}'' at ten mile intervals, and from ``\texttt{two hundred}'' to ``\texttt{a thousand miles}'' at intervals of one hundred miles.

\begin{varwidth}{0.9\textwidth}
\begin{itemize}
    \item Clean prompt: 
    \begin{itemize}
    \item ``\texttt{In the United Kingdom, <placename> is a place located near the city of}''
    \end{itemize}
\end{itemize}
\end{varwidth}

\begin{varwidth}{0.9\textwidth}
\begin{itemize}
    \item Corrupted prompt: 
    \begin{itemize}
    \item ``\texttt{In the United Kingdom, <placename> is a place located <distance> from the city of}''
    \end{itemize}
\end{itemize}
\end{varwidth}

There, however, emerges a key point of concern regarding the approach to follow in patching the activations, as any quantitative expression of distance will be composed of two or more tokens, while the corresponding ``\texttt{near}'' in the \textit{clean} prompt is composed of only one token. That creates some level of asymmetry in the patching, as the activations corresponding to ``\texttt{near the city of}'' collected from the \textit{clean} prompt substitute only part of those of the quantitative expression of distance (e.g., ``\texttt{five miles from the city of}'') rather than providing a one-to-one correspondence. We explored several avenues, including different approaches to padding and adding additional words to the \textit{clean} token (e.g., using ``\texttt{geographically near}'' instead of ``\texttt{near}''), but none seemed to provide clear advantages in our preliminary experiments, while introducing further complexity. Thus, we decided to proceed with the simplest approach, without any padding or additions for this case study.

The activation patching process generally follows the guidelines proposed by \textcite{zhang2024bestpracticesactivationpatching} using the \texttt{transformer\_lens} library\footnote{https://transformerlensorg.github.io/TransformerLens/}. The activations from the \textit{clean} prompt are substituted in the \textit{patched} computation using a five-layer sliding window at the Multi-Layer Perceptron (MLP) output hook, which seemed to be the most sensitive during our preliminary tests. As neither the \textit{clean} nor the \textit{corrupted} prompt have a necessarily unique answer, we measure the patching effect as the difference between two Kullback–Leibler (KL) divergences: the KL divergence of the next token probabilities generated by the \textit{corrupted} prompt from those generated by the \textit{clean} prompt, and the KL divergence of those generated by the \textit{patched} computation from those generated by the \textit{clean} prompt.

\section{4. Preliminary results}

\begin{figure}
    \centering
    \includegraphics[width=0.8\textwidth]{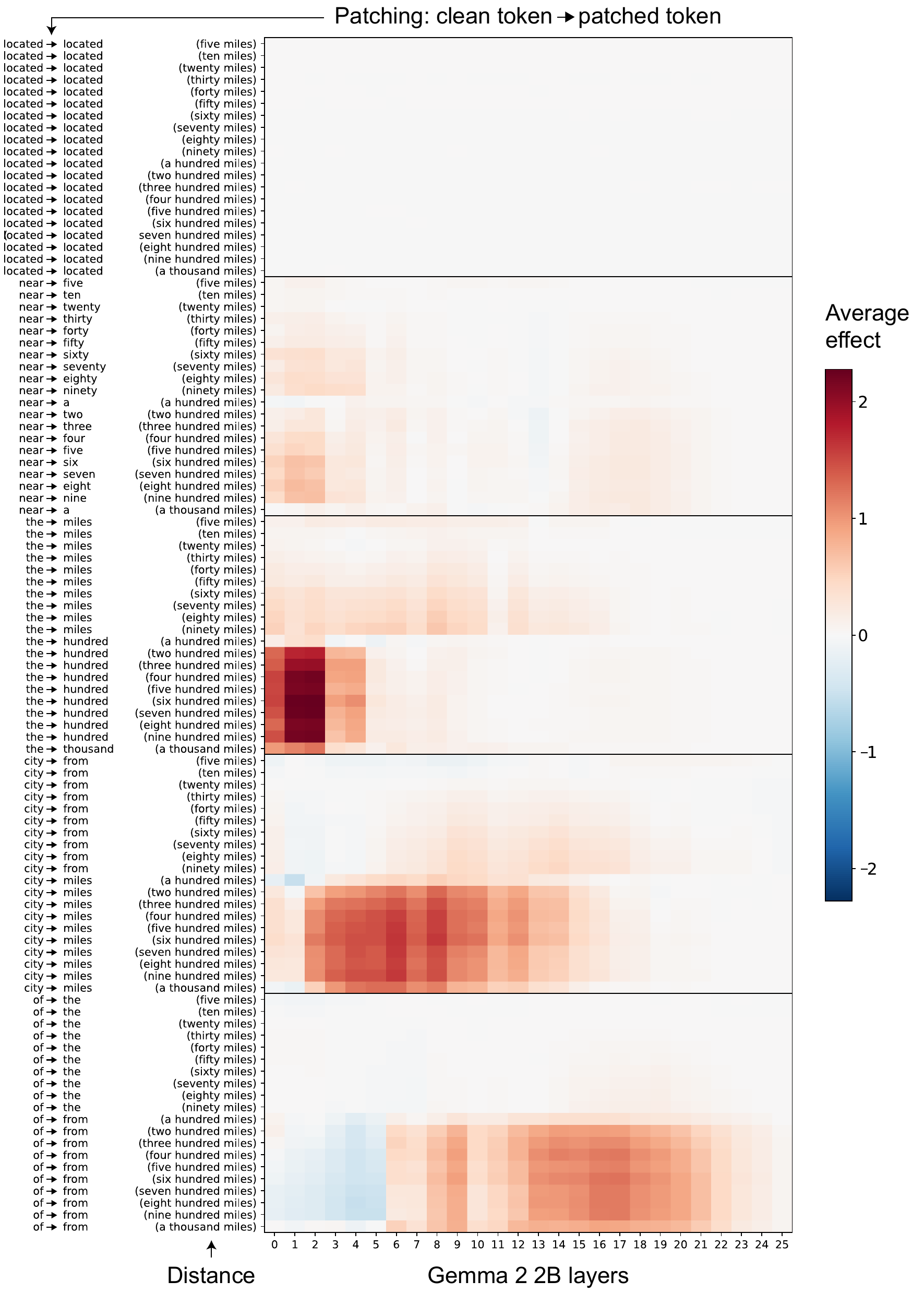}
    \caption{Average effect (colour) of patching the activations from the \textit{clean} prompt to the corresponding token in the \textit{patched} computation (vertical axis, left labels) for each distance (vertical axis, right label) and model layer (horizontal axis) across the 249 placenames included in the experiment. The tokens are patched during the computation, rather than in the prompt. Thus, the process should not be confused with a word-by-word substitution in the prompt.}
    \label{fig:patchingres}
\end{figure}

Our preliminary results provide initial insight into which parts of the LLM process relative space and reveal important aspects of how (and whether) LLMs understand geographic prompts. Figure \ref{fig:patchingres} illustrates the effect of \textit{patching} on the output: the horizontal position indicates the LLM layer being patched, while the vertical axis indicates the token and the distance that is being patched. Positive values indicate that patching the computation of the \textit{corrupted} prompt has rendered the \textit{patched} output more similar to the \textit{clean} output; values close to zero indicate that the patching has had no effect; negative values indicate that the patching has rendered the \textit{patched} output even more different from the \textit{clean} prompt output, possibly due to the disruption of core language functionalities.

The patching of the third token in the first layers of the LLM has a strong impact on the output (see the third block from the top in Figure \ref{fig:patchingres}). The third token corresponds to the end of the stated quantity in the \textit{corrupted} prompt (e.g., ``\texttt{hundred}'' in ``\texttt{located two hundred miles from the city of}''). Thus, the effect seems related to the disruption of processing of the numerical information, leading the LLM to focus on ``\texttt{near}'' rather than about the stated distance. The effect then becomes less prominent and more diffuse in the fourth and fifth tokens, which correspond to a disruption of relative geographic space information (i.e., ``\texttt{miles from}'') and seem to indicate a more complex encoding of geographic information across the neural network. It is also notable how longer distances show a much stronger effect, especially when the \textit{corrupted} prompt enters the range of hundreds of miles. There are, however, two notable exceptions, which correspond to ``\texttt{a hundred}'' and ``\texttt{a thousand miles}'', which seem to show a more tenuous effect. That seems rooted in a lower overall divergence of the probabilities generated for those distances from the probabilities generated by ``\texttt{near}'', which may be related to either a less literal interpretation of the distance or the use of a different geographic scale for \textit{near}. 

We thus observe a complex interweaving of different conceptualisations and scales in how LLMs interpret prompts. However, a broader analysis on a larger set of placenames and prompts is necessary to validate these preliminary results.

\printbibliography

\end{document}